\documentclass{article}
\usepackage{spconf,amsmath,graphicx,setspace}
\usepackage{amssymb}
\usepackage{algorithm}
\usepackage{algorithmic}
\usepackage{textcomp}
\usepackage{subfigure}
\usepackage{tipa}
\usepackage{cite} 
\usepackage{url}
\usepackage{float}
\usepackage{multirow}
\usepackage{bm}
\usepackage{svg}

\usepackage{makecell}
\usepackage{array}
\usepackage{tabularx}
\newcolumntype{C}[1]{>{\centering\arraybackslash\vspace{0pt}}m{#1}}

\title{Diversity-Driven Generative Dataset Distillation Based on \\ Diffusion Model with Self-Adaptive Memory}
\name{Mingzhuo Li$^{\dagger}$ \qquad Guang Li$^{\dagger}$ \qquad Jiafeng Mao$^{\dagger\dagger}$ \qquad Takahiro Ogawa$^{\dagger}$ \qquad Miki Haseyama$^{\dagger}$ 
\thanks{
This research was supported in part by JSPS KAKENHI Grant Numbers JP23K21676, JP24K02942, JP24K23849, and JP25K21218.
}}
\address{
$^{\dagger}$Hokkaido University \qquad \qquad
$^{\dagger\dagger}$The University of Tokyo \\
\texttt{\small{\{mingzhuo, guang, ogawa, mhaseyama\}@lmd.ist.hokudai.ac.jp}} \qquad \texttt{\small{mao@hal.t.u-tokyo.ac.jp}}
}

\begin{document}
\ninept
\maketitle
%
% \doublespacing
% \begin{spacing}{2}
\begin{abstract}
Dataset distillation enables the training of deep neural networks with comparable performance in significantly reduced time by compressing large datasets into small and representative ones. Although the introduction of generative models has made great achievements in this field, the distributions of their distilled datasets are not diverse enough to represent the original ones, leading to a decrease in downstream validation accuracy. In this paper, we present a diversity-driven generative dataset distillation method based on a diffusion model to solve this problem. We introduce self-adaptive memory to align the distribution between distilled and real datasets, assessing the representativeness. The degree of alignment leads the diffusion model to generate more diverse datasets during the distillation process. Extensive experiments show that our method outperforms existing state-of-the-art methods in most situations, proving its ability to tackle dataset distillation tasks.
\end{abstract}
\begin{keywords}
Dataset distillation, diffusion model, self-adaptive memory.
\end{keywords}

\section{Introduction}
The rapid development of deep learning leads to the widespread popularity and adoption of deep neural networks, which make full use of large amounts of data and achieve remarkable results in the field of computer vision \cite{adadi2021survey}. However, the reliance on data extends the training process to many hours or even days and imposes substantial demands on computing resources \cite{alzubaidi2021review}. Moreover, storing and maintaining such huge amounts of data is time-consuming and costly. Therefore, dataset distillation \cite{wang2018datasetdistillation} is proposed to solve these problems by distilling the knowledge of the original dataset into a new dataset, which is much smaller, yet models trained on it can achieve similar performance compared to those trained on the original dataset. 

\begin{figure}[t]
    \centering
    \subfigure[DiT \cite{Peebbles2023DiT}]{\label{grad:dit}  \includegraphics[width=0.45\linewidth]{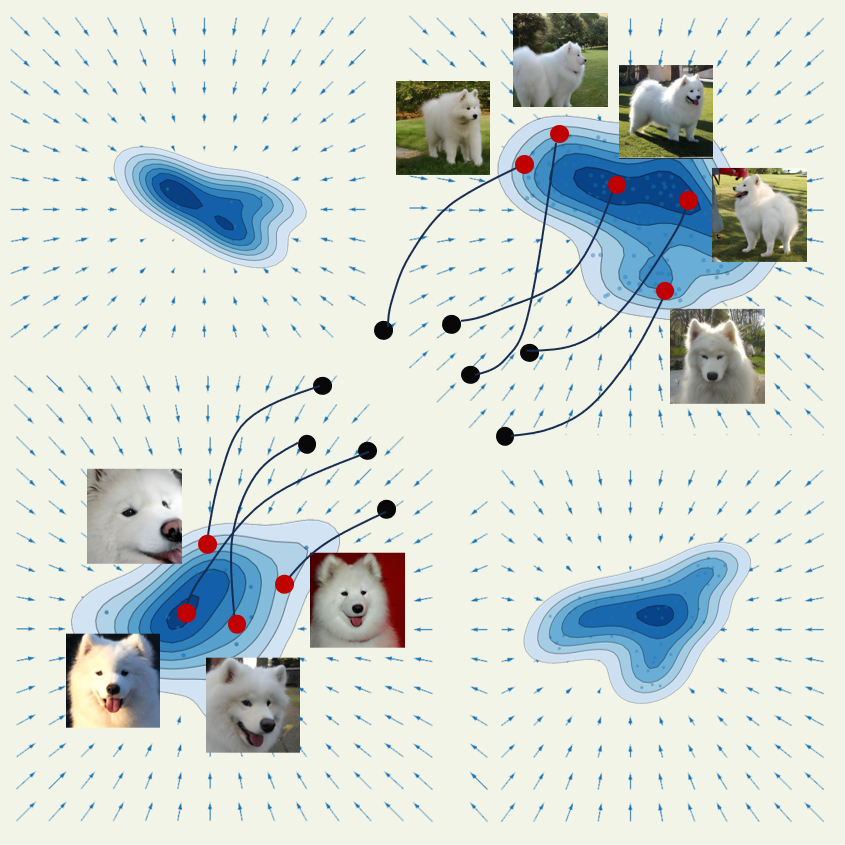}}
    \subfigure[Ours]{\label{grad:ours} \includegraphics[width=0.45\linewidth]{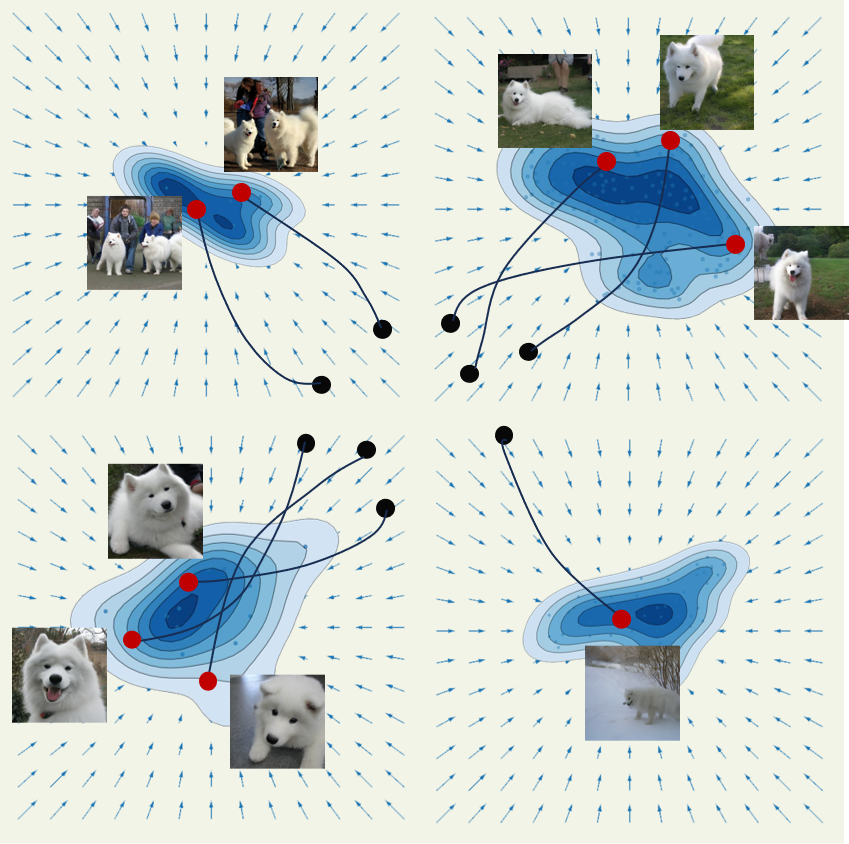}}
    \caption{A demonstration of the gradient field during the distillation. Blue areas stand for the original distribution. Arrows follow the direction of gradient descent. Black lines show the distillation process with black dots as random noise and red dots as distilled images. 
    The proposed method covers more areas, showing better diversity.
    }
    \label{grad}
\end{figure}

\begin{figure*}[t]
        \centering
        \includegraphics[width=17cm]{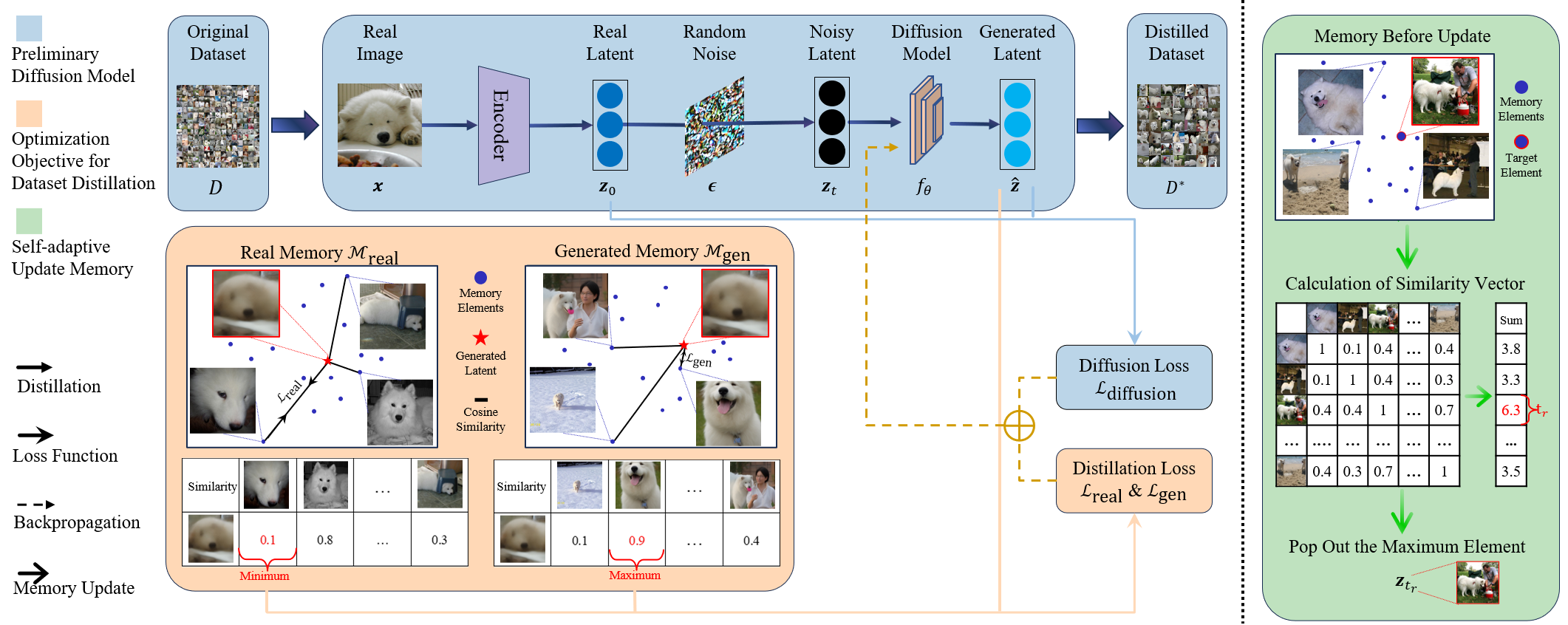}
        \caption{The distillation process of our method. Randomly selected images are input to the diffusion model to obtain the generated latents and then the diffusion loss $\mathcal{L}_\text{diffusion}$. Two memory sets consisting of real and generated latents, respectively, assist in calculating diversity loss $\mathcal{L}_\text{real}$ and $\mathcal{L}_\text{gen}$. The memory takes self-adaptive updates based on the similarity vector after each epoch.}
        \label{fig_2}
\end{figure*}

Dataset distillation has received widespread attention since it was proposed, and a substantial amount of research has contributed to its rapid development \cite{li2022awesome, yu2023review}. Nowadays, dataset distillation methods can be categorized into non-generative and generative approaches. Traditional non-generative dataset distillation methods optimize a specific number of images determined by the concept of images per class (IPC) to obtain the distilled dataset using algorithms such as gradient matching \cite{zhao2021datasetcondensation, kim2022IDC}, trajectory matching \cite{cazenavette2022MTT, li2023ddpp, li2024iadd} and kernel-based methods \cite{nguyen2021kernel1, nguyen2021kernel2}. Generative models can encode information into their structure and use it to generate synthetic images, thus offering the idea of distilling the knowledge of the original dataset into the generative model to obtain distilled datasets \cite{zhao2022synthesizing, wang2023dim, li2024generative}.
In each distillation process, traditional methods optimize one dataset of a specific size defined by IPC. Generative methods, however, obtain one model that can generate the distilled dataset of any size within a short time. 
This flexibility makes dataset distillation free from the constraints of IPC, thus saving a significant amount of time when distillation needs to be executed more than once, which is common in various downstream tasks of dataset distillation, such as continue learning \cite{yang2023continualLearn}, federated learning \cite{jia2024fedLearn}, and privacy preservation \cite{li2020soft, li2022compressed}.
\par
Among generative models, diffusion models like Imagen \cite{saharia2023imagen} and Stable Diffusion \cite{rombach2022high} become the most attractive approaches. They acquire the ability to recover an image from random noise by learning how to predict noise from noisy images, receiving broad recognition for their performance stability and distillation versatility. There are studies \cite{gu2024minimax, su2024diffusion, moser2024ld3m} trying to use diffusion models to gain better downstream validation accuracy for dataset distillation. However, there are areas containing few images in the original dataset, which are difficult for models to learn. Without effective intervention, previous works may neglect these areas, resulting in the distilled datasets not being diverse enough to represent the original ones, as shown in Fig.~\ref{grad}-\subref{grad:dit}, decreasing the downstream accuracy. To address this problem, a new method is necessary to facilitate adjustments to the distribution, assisting the diffusion model to generate more diverse distilled datasets.

\par

In this paper, we propose a diversity-driven generative dataset distillation method to obtain distilled datasets that more accurately represent the original datasets. We align the distribution between distilled and real datasets to assess the representativeness. Considering that the original dataset is excessively large for alignment, auxiliary memory is introduced to substitute the original dataset, reducing the amount of data to be processed. The alignment degree is used for model optimization during the distillation process, leading the diffusion model to generate datasets of higher diversity. At the same time, the memory incorporates self-adaptive updates by popping out images that are most similar to others, maintaining its size for computation while remaining eligible to replace the original dataset. With such improvement, datasets generated by our method exhibit better diversity to represent the original ones as shown in Fig.~\ref{grad}-\subref{grad:ours}, thus obtaining better downstream validation accuracy. Extensive experiments have been conducted to show that our methods outperform state-of-the-art methods in most situations with different datasets and IPC settings.
\par
The contributions of this paper can be summarized as follows:
\begin{itemize}
    \item We propose a diversity-driven generative dataset distillation method that introduces self-adaptive memory to quantify and back-propagate the diversity during the distillation process, thus assisting the diffusion model in generating datasets of better diversity.
    \item We conduct extensive experiments with different IPC and dataset settings, proving that our methods have better downstream validation accuracy than state-of-the-art methods in most situations.
\end{itemize}

\section{Diversity-driven dataset distillation}
The proposed method aims to obtain distilled datasets that better represent the original ones by using diversity-driven optimization with the assistance of auxiliary memory. Our method mainly includes three parts: preliminary diffusion model, optimization objectives for dataset distillation, and self-adaptive memory. The whole distillation process is illustrated in Fig.~\ref{fig_2} and Alg.~\ref{alg1}.
\par
\subsection{Preliminary}
Diffusion models acquire the ability to recover an image from random noise by learning how to predict noise from noisy images. Latent diffusion models (LDMs), which operate on latent vectors rather than directly on image pixels, demonstrate enhanced efficacy and capability, especially on abstract and semantic information. 
\par
Here we use VAE \cite{kingma2013VAE} as the encoder and DiT \cite{Peebbles2023DiT} as the baseline model, with the diffusion process shown in the blue areas in Fig.~\ref{fig_2}. Given an original dataset $D$, the training process starts with randomly selecting a training image $\bm{x}$, which is encoded to get the real Latent $z_0$. Then random noise $\epsilon \in \mathcal{N}(\bm{0}, \bm{I})$ is added to $\bm{z}_0$ for $t$ times to get the noisy latent $\bm{z}_t$ as follows:
\begin{equation}
    \bm{z_t} = \sqrt{\overline{\alpha}_t} E(\bm{x}) + \sqrt{1 - \overline{\alpha}_t} \epsilon, 
\end{equation}
where $\overline{\alpha}_t$ is a hyper-parameter known as variance schedule and $E(\text{·})$ is the VAE encoder. We use another class encoder to obtain the conditioning vector $\bm{c}$ to construct the noise predictor of the diffusion model $\epsilon_{\theta}(\bm{z_t}, t, \bm{c})$. The diffusion model is trained by minimizing the predicted noise and the ground truth $\epsilon$ as follows:
\begin{equation}
    \mathcal{L}_{\text{diffusion}} = {||\epsilon_{\theta}(\bm{z_t}, t, \bm{c}) - \epsilon||}_2^2. 
\end{equation}
After training, images are generated by the diffusion model to construct the distilled dataset.
\par
The diffusion model aims to generate images that resemble the original ones, which is insufficient for dataset distillation. To achieve comparable accuracy, the distilled dataset should exhibit sufficient diversity to represent the original dataset, necessitating the development of robust optimization objectives.
\vspace{-5pt}

\subsection{Optimization Objective}

We simplify the diversity as the similarity between the distilled and original distribution, which leads to a naive objective of aligning the generated and real latents as follows: 
\begin{equation}
    \mathcal{L}_\text{diversity} = \arg \max_{\theta} \sum_{i=0}^{N} \sigma(\bm{\hat{z}}_{\theta}(\bm{z}_t, \bm{c}), \bm{z}_i), 
\end{equation}
where $\sigma(\text{·},\text{·})$ is the cosine similarity, $\bm{\hat{z}}_{\theta}(\bm{z}_t, \bm{c})$ is the latent of the image generated by the diffusion model $f_{\theta}$, $\bm{z}_i$ is the latent of the \textit{i}-th real image, and $N$ is the size of the original dataset. 
However, the excessive size of $N$ makes such alignment impractical.
Simply using the mini-batch for replacement tends to draw the distilled latents toward the center of the real distribution, limiting the diversity. 

\par

Minimax \cite{gu2024minimax} provides a solution to this problem by introducing the minimax diffusion criteria to obtain optimization objectives focused on dataset distillation. As shown in the orange areas in Fig.~\ref{fig_2}, With the assistance of real memory $\mathcal{M}_\text{real}$ and generated memory $\mathcal{M}_\text{gen}$, it introduces two optimization objectives $\mathcal{L}_\text{real}$ and $\mathcal{L}_\text{gen}$ to enhance the representativeness and diversity of the distilled dataset.
$\mathcal{M}_\text{real}=\{\bm{z}_r\}_{r=0}^{N_{R}}$ contains image latents from the original dataset and $\mathcal{M}_\text{gen}=\{\bm{\hat{z}}_{\theta}(\bm{z}_g, \bm{c})\}_{g=0}^{N_{G}}$ contains image latents generated by the diffusion model during the distillation process. In each epoch, cosine similarity between the generated latent and all the elements in the memory is calculated. The representativeness objective pulls close the least similar pairs in the real memory as follows: 
\begin{equation}
    \mathcal{L}_\text{real} = \arg \max_{\theta} \min_{r \in [N_{R}]} \sigma(\bm{\hat{z}}_{\theta}(\bm{z}_t, \bm{c}), \bm{z}_{r}).
\end{equation}
The diversity objective pulls away the most similar pairs in the generative memory with maximum similarity as follows:
\begin{equation}
    \mathcal{L}_\text{gen} = \arg \min_{\theta} \max_{g \in [N_{G}]} \sigma(\bm{\hat{z}}_{\theta}(\bm{z}_t, \bm{c}), \bm{\hat{z}}_{\theta}(\bm{z}_g, \bm{c})).
\end{equation}
The combination of the objective of distillation and the objective and diffusion model leads to the final optimization objective as follows: 
\begin{equation}
    \label{equ_loss}
    \mathcal{L} = \mathcal{L}_\text{diffusion} + \lambda_{r}\mathcal{L}_\text{real} + \lambda_{g}\mathcal{L}_\text{gen}, 
\end{equation}
where $\lambda_{r}$ and $\lambda_{g}$ are hyper-parameters for balancing the effects. 
The real loss $\mathcal{L}_\text{real}$ drives the distilled distribution to fit the original distribution, improving its potential to represent the original dataset. While the generative loss $\mathcal{L}_\text{gen}$ makes distilled images distribute more diversely, avoiding the possible decline in accuracy caused by images gathering in certain areas. The combination of $\mathcal{L}_\text{real}$ and $\mathcal{L}_\text{gen}$ yields mutual effects in generating diverse distilled datasets that fit the original distribution.
\vspace{-5pt}

\begin{algorithm}[t]
    \caption{Diversity-Driven Dataset Distillation}
    \label{alg1}
    \begin{algorithmic}[1]
    \REQUIRE 
    $f_{\theta}$: a pre-trained diffusion model parameterized by $\theta$;
    $\mathcal{D}=\{(\bm{x}, y)\}$: the original dataset;
    $E$: the VAE encoder;
    $E_c$: the class encoder;
    $\mathcal{M}_\text{real}$: the memory of real latents;
    $N_R$: the pre-defined size of real memory;
    $N_r$: current size of real memory;
    $\mathcal{M}_\text{gen}$: the memory of generated latents;
    $N_g$: current size of generative memory;
    $N_G$: the pre-defined size of generative memory;
    $\bm{\epsilon}$: random Gaussian noise
    \ENSURE
    $f_{\theta^{\ast}}$: the fine-tuned diffusion model; 
    $\mathcal{D}^{\ast}$: the distilled dataset
    
    \FOR{ each epoch $i$ = 1 to I }
    \STATE
    Obtain the real latent: $\bm{z}_i=E(\bm{x}_i)$
    \STATE
    Obtain the class latent: $\bm{c}=E_c(y)$
    \STATE
    Obtain the generated latent: $\bm{\hat{z}}_{\theta}(\bm{z}_{i}, \bm{c})$
    \STATE
    Update the model parameter with Eq.~(\ref{equ_loss})
    
    \STATE
    Enqueue the real latent: $\mathcal{M}_\text{real} \xleftarrow{} \bm{z}_i$
    \WHILE{$N_r > N_{R}$}
    \STATE
    Locate the index of the target latent $t_r$ with Eq.~(\ref{s_real})
    \STATE
    Pop out the target latent: $\bm{z}_{t_r}$
    \ENDWHILE
    
    \STATE
    Enqueue the generated latent: $\mathcal{M}_\text{gen} \xleftarrow{} \bm{\hat{z}}_{\theta}(\bm{z}_{i}, \bm{c})$
    \WHILE{$N_g > N_{G}$}
    \STATE
    Locate the index of the target latent $t_g$ with Eq.~(\ref{s_gen})
    \STATE
    Pop out the target latent $\bm{\hat{z}}_{\theta}(\bm{z}_{t_g}, \bm{c})$
    \ENDWHILE
    
    \ENDFOR
    \STATE
    Save the fine-tuned diffusion model $f_{\theta^{\ast}}$
    \STATE
    Generate the distilled dataset :$\mathcal{D}^{\ast} = f_{\theta^{\ast}}(\bm{\epsilon}, \bm{c})$
    \end{algorithmic}
\end{algorithm}

\subsection{Self-adaptive Memory}

 Due to the consideration of storage burden, the memory size is limited, resulting in the replacement of memory elements during the distillation process. Minimax pops out the oldest ones for both real and generative memory, neglecting the effect on memory distribution, which causes its distribution to shift towards a random subset of the original dataset because of random sampling. Considering that the target of the memory is to encapsulate the entire dataset within a reduced size, which requires it to be diverse enough to represent the original distribution, this neglect may interfere with the operation of the optimization objectives.

\par

Therefore, we introduce self-adaptive updates for the memory to ensure diversity with the assistance of similarity vectors for real memory and generative memory, respectively. When the real memory's current size $N_r$ extends the pre-defined size $N_R$, the update starts with the calculation of the similarity vector. For each element in the memory, the summary of its cosine similarity with all the elements is calculated. The similarity vector is obtained by including the summary values of all the elements. The index of maximum similarity degree in the vector $t_r$ is calculated as follows:
\begin{equation}
\label{s_real}
    t_r = \arg \max_{i \in [N_r]} \sum_{j=0}^{N_r} \sigma(\bm{z}_i, \bm{z}_{j}).
\end{equation}

\vspace{-2mm}
Then the corresponding latent $\bm{z}_{t_r}$ is located and popped out to keep the memory diverse to represent the original distribution. The update of real memory is illustrated in the green areas in Fig.~\ref{fig_2}.

\par

Similarly, when the generative memory's current size $N_g$ extends the pre-defined size $N_G$, the index $t_g$ is calculated as follows: 
\begin{equation}
\label{s_gen}
    t_g = \arg \max_{i \in [N_g]} \sum_{j=0}^{N_g} \sigma(\bm{\hat{z}}_{\theta}(\bm{z}_i, \bm{c}), \bm{\hat{z}}_{\theta}(\bm{z}_j, \bm{c})).
\end{equation}
The corresponding latent $\bm{\hat{z}}_{\theta}(\bm{z}_{t_g}, \bm{c})$ is then popped out.
For cosine similarity, higher scores indicate greater similarity between elements. Therefore, latents with higher similarity degrees, which may overly cluster the distribution when integrated, are prioritized for removal to maintain diversity. Guided by this principle, we iteratively remove the latent with the highest similarity score until the memory size meets the predefined limit, ensuring the remaining memory elements are more diverse and better represent the overall distribution. By employing this self-adaptive update strategy, the memory progressively achieves a balanced representation of diversity and accuracy, mitigating the risk of performance degradation caused by insufficient representativeness. 

\begin{table*}
    \centering
    \footnotesize
    % \arraystretch{3}
    \renewcommand{\arraystretch}{1.5}
    
    \tabcolsep=5pt
    \caption{Comparison of downstream validation accuracy with other SOTA methods on ImageWoof. The best results are marked in bold.}
    \label{tab_main}
    \begin{tabular}{C{37pt}| C{45pt}| C{28pt} C{45pt} C{42pt}| C{28pt} C{28pt} C{32pt} C{45pt} C{34pt}| C{39pt}}
        \hline
        IPC (Ratio) & Test Model & Random & K-Center\cite{sener2017kcenter} & Herding \cite{Welling2009Herding} & DiT \cite{Peebbles2023DiT} & DM \cite{zhao2023DM} & IDC-1 \cite{kim2022IDC} & Minimax \cite{gu2024minimax} & Ours & Full Dataset \\ 
        \hline
         
        & ConvNet-6 & $24.3_{\pm 1.1}$ & $19.4_{\pm 0.9}$ & $26.7_{\pm 0.5}$ & $34.2_{\pm 1.1}$ & $26.9_{\pm 1.2}$ & $33.3_{\pm 1.1}$ & $34.3_{\pm 0.5}$ & \bm{$36.2_{\pm 0.3}$} & $86.4_{\pm 0.2}$ 
        \\ 
        10 (0.8\%) & ResNetAP-10 & $29.4_{\pm 0.8}$ & $22.1_{\pm 0.1}$ & $32.0_{\pm 0.3}$ & $34.7_{\pm 0.5}$ & $30.3_{\pm 1.2}$ & $37.3_{\pm 0.4}$ & $35.7_{\pm 0.3}$ & \bm{$38.1_{\pm 0.3}$} & $87.5_{\pm 0.5}$ 
        \\
        & ResNet-18 & $27.7_{\pm 0.9}$  & $21.1_{\pm 0.4}$ & $30.2_{\pm 1.2}$ & $34.7_{\pm 0.4}$ & $33.4_{\pm 0.7}$ & $36.9_{\pm 0.4}$ & $35.5_{\pm 0.7}$ & \bm{$37.5_{\pm 0.6}$} & $89.3_{\pm 1.2}$ 
        \\ 
        \hline
         
        & ConvNet-6 & $29.1_{\pm 0.7}$ & $21.5_{\pm 0.8}$ & $29.5_{\pm 0.3}$ & $36.1_{\pm 0.8}$ & $29.9_{\pm 1.0}$ & $35.5_{\pm 0.8}$ & $36.3_{\pm 0.4}$ & \bm{$37.7_{\pm 0.7}$} & $86.4_{\pm 0.2}$ 
        \\ 
        20 (1.6\%) & ResNetAP-10 & $32.7_{\pm 0.4}$ & $25.1_{\pm 0.7}$ & $34.9_{\pm 0.1}$ & $41.1_{\pm 0.8}$& $35.2_{\pm 0.6}$ & $42.0_{\pm 0.4}$ & $43.9_{\pm 0.7}$ & \bm{$44.5_{\pm 0.8}$} & $87.5_{\pm 0.5}$ 
        \\ 
        & ResNet-18 & $29.7_{\pm 0.5}$ & $23.6_{\pm 0.3}$ & $32.2_{\pm 0.6}$ & $40.5_{\pm 0.5}$ & $29.8_{\pm 1.7}$ & $38.6_{\pm 0.2}$ & $41.1_{\pm 1.0}$ & \bm{$42.2_{\pm 0.5}$} & $89.3_{\pm 1.2}$ 
        \\ 
        \hline
         
        & ConvNet-6 & $41.3_{\pm 0.6}$ & $36.5_{\pm 1.0}$ & $40.3_{\pm 0.7}$ & $46.5_{\pm 0.8}$ & $44.4_{\pm 1.0}$ & $43.9_{\pm 1.2}$ & $51.3_{\pm 0.8}$ & \bm{$52.7_{\pm 1.0}$} & $86.4_{\pm 0.2}$ 
        \\ 
        50 (3.8\%) & ResNetAP-10 & $47.2_{\pm 1.3}$ & $40.6_{\pm 0.4}$ & $49.1_{\pm 0.7}$ & $49.3_{\pm 0.2}$ & $47.1_{\pm 1.1}$ & $48.3_{\pm 1.0}$ & $55.1_{\pm 0.7}$ & \bm{$56.9_{\pm 0.2}$} & $87.5_{\pm 0.5}$ 
        \\ 
        & ResNet-18 & $47.9_{\pm 1.8}$ & $39.6_{\pm 1.0}$ & $48.3_{\pm 1.2}$ & $50.1_{\pm 0.5}$ & $46.2_{\pm 0.6}$ & $48.3_{\pm 0.8}$ & $54.3_{\pm 1.1}$ & \bm{$56.0_{\pm 0.6}$} & $89.3_{\pm 1.2}$ 
        \\
        
        \hline
    \end{tabular}
\end{table*}

\begin{figure*}[t]
        \centering
        \includegraphics[width=17cm]{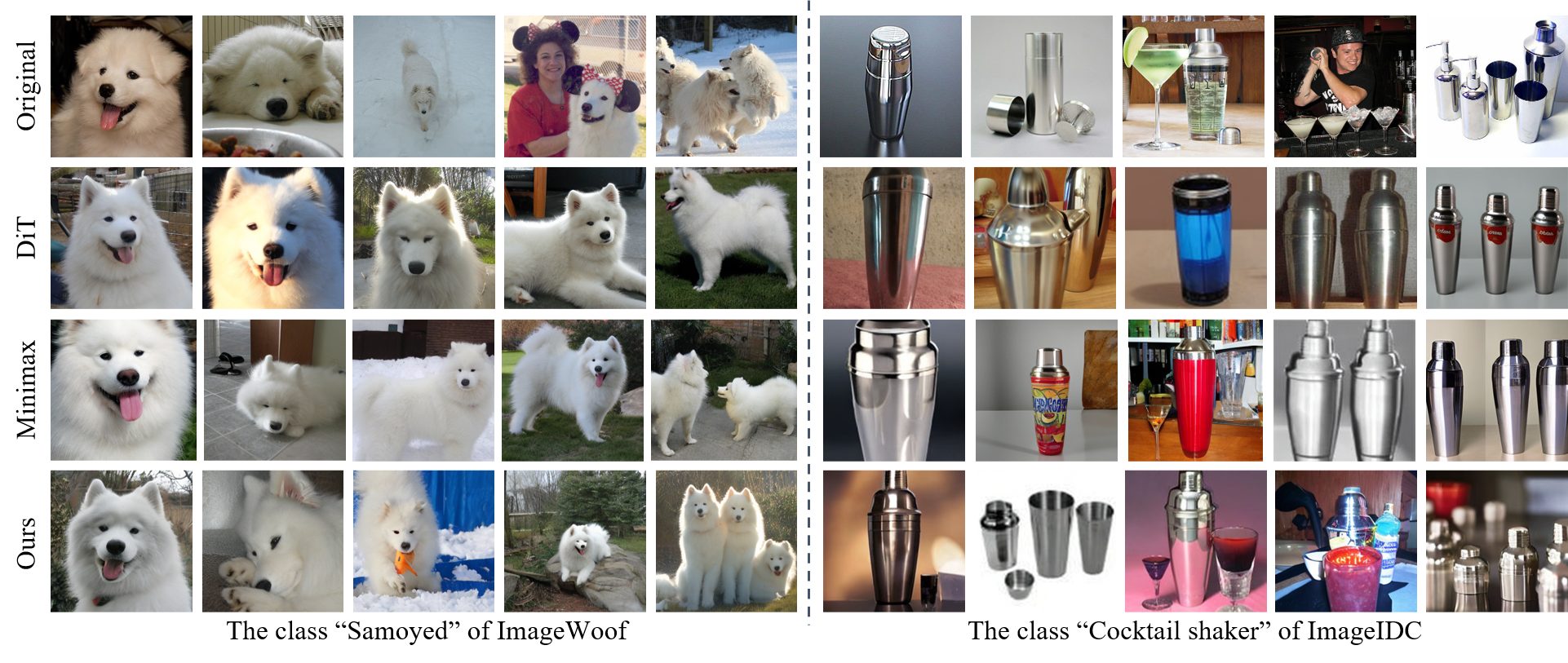}
        \caption{Comparison of visualization results between original images and distilled images of DiT, Minimax, and our method with IPC = 50.} 
        \label{fig_vi}
\end{figure*}

% \vspace{-10mm}
\section{Experiments}
\begin{figure*}[t]
    \centering
    \subfigure[]{\label{hyper:real}  \includegraphics[width=5.8cm]{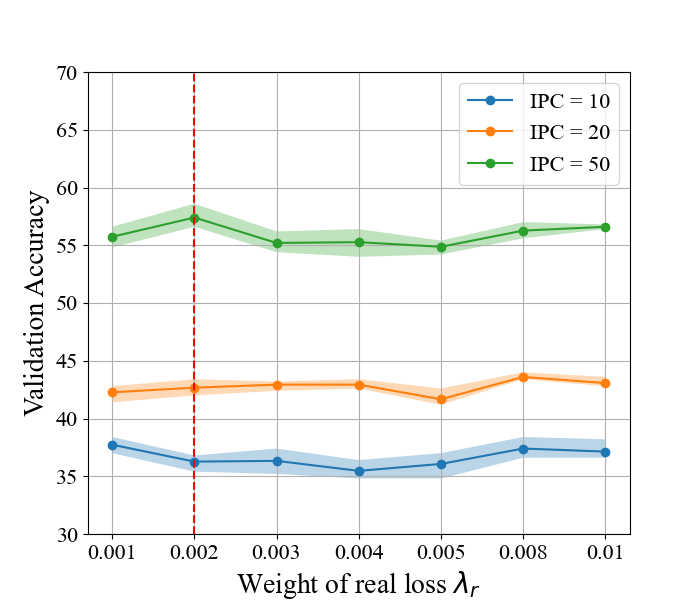}}
    \subfigure[]{\label{hyper:gen} \includegraphics[width=5.8cm]{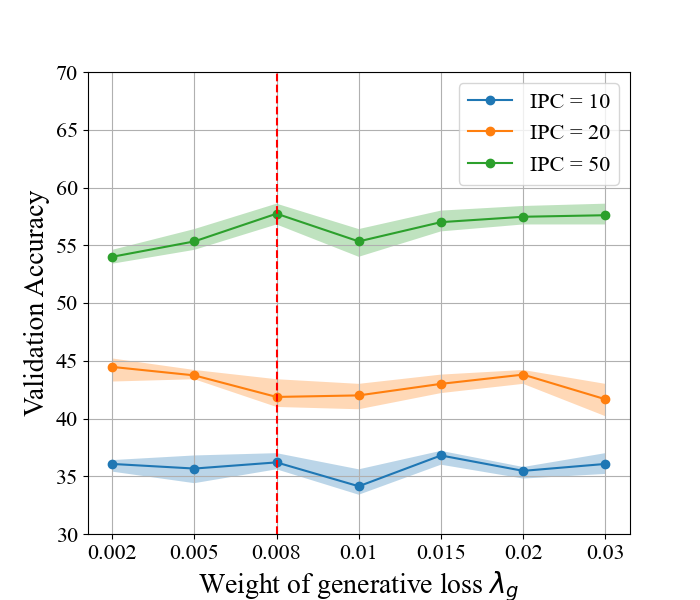}}
    \subfigure[]{\label{hyper:size}  \includegraphics[width=5.8cm]{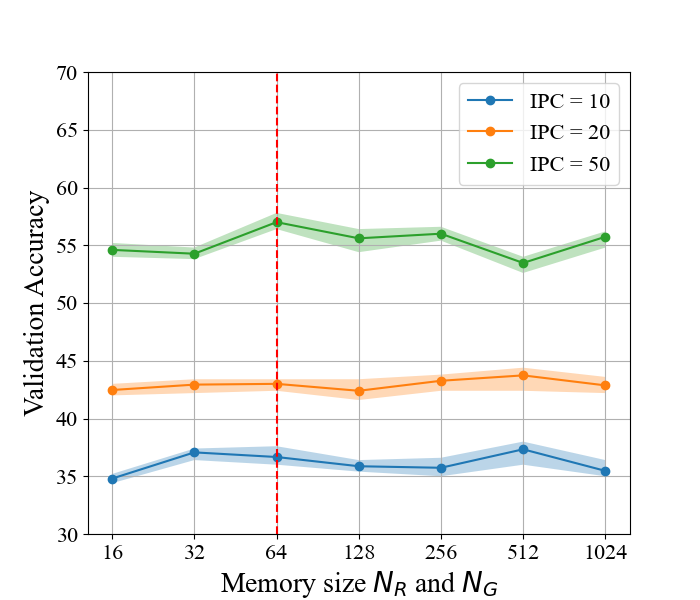}}
    \caption{Analysis on hyperparameters of weight of real loss $\lambda_r$, weight of generative loss $\lambda_g$, and memory size $N_R$ and $N_G$. The results are obtained with ResNetAP-10 on ImageWoof with different IPC settings. Each setting is conducted 3 times, with points showing the average values and shadows covering the areas between minimum and maximum accuracy. The performance of the settings adopted in the primary experiments is marked with the dashed lines.}
    \label{hyper}
\end{figure*}

\subsection{Datasets and Evaluation}
Extensive experiments have been conducted to verify the effectiveness of the proposed method on three benchmark subsets of the full-sized ImageNet \cite{jia2009imageNet} dataset: ImageWoof \cite{fastai20imageNettte}, ImageNette \cite{fastai20imageNettte}, and ImageIDC \cite{kim2022IDC}. ImageWoof contains 10 classes of specific dog breeds, a challenging dataset for classification. ImageNette contains 10 specific classes that are easy to classify. ImageIDC contains 10 classes randomly selected from ImageNet. For evaluation, we compare the downstream validation accuracy with several dataset distillation methods, including baseline methods DiT \cite{Peebbles2023DiT}, along with traditional dataset selection methods like Random, K-Center \cite{sener2017kcenter}, and Herding \cite{Welling2009Herding}. We also evaluate our method against state-of-the-art (SOTA) methods like DM \cite{zhao2023DM}, IDC-1 \cite{kim2022IDC}, and Minimax \cite{gu2024minimax}. The downstream models for validation include ConvNet-6 \cite{gidaris2018convNet}, ResNet-18 \cite{he2016resNetAP}, and ResNet-10 with average pooling (ResNetAP-10) \cite{he2016resNetAP}, with a learning rate of 0.01 and top-1 accuracy being reported.

\par

For the diffusion model, a pre-trained DiT \cite{Peebbles2023DiT} with Difffit \cite{xie2023Difffit} for fine-tuning is adopted as the baseline model, and VAE \cite{kingma2013VAE} as the encoder. The input image is randomly arranged and transformed to 256 $\times$ 256 pixels. The number of denoising steps in the sampling process is 50. The distillation process lasts for 8 epochs with a mini-batch size of 8. An AdamW with a learning rate of 1e-3 is adopted as the optimizer. The memory size is 64. The weighting hyper-parameters $\lambda_{r}$ and $\lambda_{g}$ in Eq.~(\ref{equ_loss}) are set as 0.002 and 0.008, respectively. Each experiment is repeated 3 times, and the mean value and standard deviation are recorded. 
% All the experiments are conducted on a single RTX A6000 GPU.

\subsection{Benchmark Results}
We compare our method in the most challenging ImageWoof dataset with different classification models and various IPC settings. As shown in Table~\ref{tab_main}, our method demonstrates superior accuracy across all settings, especially in low IPC settings, proving the method's ability to improve the performance of dataset distillation. 

\par

Then, we discuss the generalization performance of the proposed method by conducting experiments on various datasets. As shown in Table~\ref{tab_dataset}, the performance trend on ImageNette and ImageIDC generally corresponds with that on ImageWoof, except that the Random method gets the best accuracy in large IPC settings on ImageNette. This is because the classes in the ImageNette are easy to discriminate, diminishing the effect of distillation. As the difficulty increases on ImageIDC, the impact of distillation begins to take effect, and our method demonstrates superior accuracy compared to others.

\par

We also compare the visualization results between our method and other SOTA methods in Fig.~\ref{fig_vi}. SOTA methods obtain realistic images with high-quality appearance, but the images tend to exhibit similarity in poses and a lack of details, indicating a limitation in diversity. 
Our method improves both the diversity of the distilled distribution and the coverage of the distilled dataset, resulting in datasets with more detailed and varied images.

\subsection{Ablation Study}
\label{ablation}
In updating memory sets, we hypothesize that the maximum strategy that removes the latent with the maximum similarity degree helps improve the effectiveness of both the real and generated memory. We validate the hypothesis in Table~\ref{tab_ablation}, where we conduct experiments with different update strategies. Results show that the combination of the maximum strategy for the real memory and the maximum strategy for the generative memory achieves the best accuracy in all IPC settings. The maximum strategy improves the real memory's diversity, but results in different effects on performance for the two memories. 
Improving the real memory's diversity leads to a distilled distribution that better represents the original distribution. However, the distilled datasets may achieve worse accuracy without the ability to cover the distribution, causing a reduction of diversity due to the concentration of images in specific areas. Improving the generative memory's diversity promotes a sparse distribution of the distilled dataset, thereby improving distribution coverage and enhancing diversity. As a result, combining maximum strategy and maximum strategy effectively leverages the benefits derived from enhanced memory diversity, achieving the best performance among different combinations.

\begin{table}
    \centering
    \footnotesize
    \renewcommand{\arraystretch}{1.5}
    \caption{Comparison of downstream validation accuracy with other SOTA methods on different ImageNet subsets. The results are obtained with ResNetAP-10. The best results are marked in bold.}
    \label{tab_dataset}
    \begin{tabular*}{\linewidth}{c|c|cccc}
        \hline
        & IPC & Random & DiT \cite{Peebbles2023DiT} & Minimax \cite{gu2024minimax} & Ours \\

        \hline
        \multirow{3}*{\rotatebox{90}{ImageNette}} & 10 & $54.2_{\pm 1.6}$  & $59.1_{\pm 0.7}$ & $59.8_{\pm 0.3}$ & \bm{$60.5_{\pm 0.3}$} \\
        & 20 & $63.5_{\pm 0.5}$ & \bm{$64.8_{\pm 1.2}$} & $64.1_{\pm 0.4}$ & $64.5_{\pm 0.4}$ \\
        & 50 & \bm{$76.1_{\pm 1.1}$} & $73.3_{\pm 0.9}$ & $75.2_{\pm 0.2}$ & $75.5_{\pm 0.3}$ \\

        \hline
        \multirow{3}*{\rotatebox{90}{ImageIDC}} & 10  & $48.1_{\pm 0.8}$ & $54.1_{\pm 0.4}$ & $46.9_{\pm 0.6}$ & \bm{$54.7_{\pm 0.4}$}
        \\
        & 20 & $52.5_{\pm 0.9}$ & $58.9_{\pm 0.2}$ & $58.2_{\pm 0.7}$ & \bm{$63.0_{\pm 0.4}$}
        \\
        & 50 & $68.1_{\pm 0.7}$ & $64.3_{\pm 0.6}$ & $67.1_{\pm 0.4}$ & \bm{$69.3_{\pm 0.2}$}
        \\
        \hline
         
    \end{tabular*}
\end{table}

\begin{table}
    \centering
    \footnotesize
    \caption{Comparison of downstream validation accuracy for different update strategies of memory sets. The results are obtained with ResNetAP-10 on ImageWoof. The best results are marked in bold.}
    \label{tab_ablation}
    \centering
    \begin{tabular}{cc|ccc}
        \hline
        $\mathcal{M}_\text{real}$ & $\mathcal{M}_\text{gen}$ & IPC = 10 & IPC = 20 & IPC = 50
        \\
        \hline
        min & min & $23.9_{\pm 0.3}$ & $31.5_{\pm 0.3}$ & $36.3_{\pm 0.9}$
        \\
        max & min & $23.9_{\pm 0.1}$ & $29.7_{\pm 0.3}$ & $35.5_{\pm 0.5}$
        \\
        min & max & $34.4_{\pm 0.4}$ & $41.7_{\pm 0.2}$ & $55.4_{\pm 0.4}$
        \\
        max & max & \bm{$37.7_{\pm 0.4}$} & \bm{$44.5_{\pm 0.8}$} & \bm{$56.9_{\pm 0.3}$}
        \\
        \hline
    \end{tabular}
\end{table}

\subsection{Analysis on Hyperparameters}
We conduct several experiments to evaluate the effects of hyperparameters. As shown in Fig.~\ref{hyper}-\subref{hyper:real}, the overall performance shows a smooth curve for the weight of real loss $\lambda_r$. For the weight of generative loss $\lambda_g$, as shown in Fig.~\ref{hyper}-\subref{hyper:gen}, the increase in $\lambda_g$ within a certain range improves the validation accuracy gradually, especially in large IPC settings. For the memory size $N_R$ and $N_G$, the self-adaptive update approach enables the memory to represent the distribution even with a small size, making the accuracy insensitive to changes in memory size as shown in Fig.~\ref{hyper}-\subref{hyper:size}. Based on the performance across all IPC settings, we set $\lambda_r$ as 0.002, $\lambda_g$ as 0.008, and $N_R$ and $N_G$ as 64 for the primary experiments. 
The method demonstrates stability, as its performance remains consistent across various settings.

\section{Conclusion}
In this paper, we have proposed a diversity-driven generative dataset distillation method based on a diffusion model. A key innovation of our method is the introduction of two self-adaptive memory sets for evaluating and improving the diversity of the distilled dataset, aiming to enhance its ability to represent the original dataset. Self-adaptive updates are conducted based on the similarity vectors to ensure the stability and effectiveness of the memory. With such efforts, the proposed method achieves state-of-the-art performance in most experiments concerning different IPC settings in various datasets. 
It verifies the effectiveness of diversity-based optimization and its ability to improve the performance of dataset distillation on downstream tasks.

\clearpage

\bibliographystyle{IEEEbib}
\bibliography{refs}

\end{document}